\documentclass{article}

\usepackage{microtype}
\usepackage{graphicx}
\usepackage{subfigure}
\usepackage{booktabs} 

\usepackage{hyperref}

\usepackage{amssymb}
\usepackage{amsmath}
\usepackage{bbm}
\usepackage{mathtools}
\usepackage{tikz}
\usepackage{xcolor}
\usepackage{pgfplots}
\pgfplotsset{compat=1.16}
\usetikzlibrary{shapes, calc, shapes, arrows}

\newcommand\independent{\protect\mathpalette{\protect\independenT}{\perp}}
\def\independenT#1#2{\mathrel{\rlap{$#1#2$}\mkern2mu{#1#2}}}

\def\X{\mathbf X}

\def\x{\mathbf x}

\def \real{\rm I\!R}

\DeclarePairedDelimiter{\ceil}{\lceil}{\rceil}
\DeclareMathOperator*{\argmin}{argmin~} 

\usepackage[accepted]{icml2020}

\icmltitlerunning{Adapting Neural Networks for Uplift Models}

\begin{document}

\twocolumn[
\icmltitle{Adapting Neural Networks for Uplift Models}




\begin{icmlauthorlist}
\icmlauthor{Mouloud Belbahri}{goo,to}
\icmlauthor{Olivier Gandouet}{goo}
\icmlauthor{Ghaith Kazma}{goo}
\end{icmlauthorlist}

\icmlaffiliation{to}{Department of Mathematics and Statistics, University of Montreal, Montreal, Canada}
\icmlaffiliation{goo}{Analytics and Modeling - Advanced Projects, TD Insurance, Montreal, Canada}

\icmlcorrespondingauthor{Mouloud Belbahri}{mouloud.belbahri@tdassurance.com}
\icmlcorrespondingauthor{Olivier Gandouet}{olivier.gandouet@tdassurance.com}

\icmlkeywords{Machine Learning, ICML}

\vskip 0.3in
]



\printAffiliationsAndNotice{}  

\begin{abstract}
\textit{Uplift} is a particular case of individual treatment effect modeling. Such models deal with cause-and-effect inference for a specific factor, such as a marketing intervention. In practice, these models are built on customer data who purchased products or services to improve product marketing. Uplift is estimated using  either i) \textit{conditional mean regression} or ii) \textit{transformed outcome regression}. Most existing approaches are adaptations of classification and regression trees for the uplift case. However, in practice, these conventional approaches are prone to overfitting. Here we propose a new method using neural networks. This representation allows to jointly optimize the difference in conditional means and the transformed outcome losses. As a consequence, the model not only estimates the uplift, but also ensures consistency in predicting the outcome. We focus on fully randomized experiments, which is the case of our data. We show our proposed method improves the state-of-the-art on synthetic and real data.

\end{abstract}

\section{Introduction}
\label{introduction}

\textit{Uplift} is a particular case of causal inference and follows the potential outcomes framework \citep{rubin1974estimating, holland1986statistics}. Uplift models are developed for randomized controlled trials (RCTs) settings where both the treatment and outcome are binary random variables. Uplift refers to modeling behavioral change that results directly from a specific treatment, such as a marketing intervention. 

Mainly known in the marketing research literature, uplift models \citep{radcliffe1999differential, hansotia2002direct, lo2002true, radcliffe2007using} provide a solution to the problem of isolating the marketing effect. Instead of modeling the positive response probability, uplift attempts to model the difference between conditional probabilities in the treatment and control groups. Uplift definition is equivalent to the one of individual treatment effect (ITE). However, uplift models performance metrics diverge from those used for ITE estimation \citep{radcliffe2011real}. 

In practice, the predicted ITE is used to rank customers in the order that they should be targeted to maximize a future marketing campaign success. To do so, an uplift model is built using a historical list of clients to whom a business sold products and services. This is often achieved in RCTs settings.
First, a pilot study is conducted. In this case, the treatment assignment (e.g. a courtesy call) is controlled, and thus the distributions of treatment and control groups are known. The model is then built on the collected data.

The customers can be segmented along two dimensions in function of the potential outcomes and the treatment, giving rise to the following groups \citep{kane2014mining} : a \textit{persuadable} client is an individual who purchases the product only if he receives a call; a \textit{sure thing} client is an individual who purchases the product whether he receives the call or not; a \textit{lost cause} client is an individual who will not purchase the product regardless of whether he receives the call or not; a \textit{do-no-disturb} client is an individual who will not purchase the product only if he receives a call.

In general, the interesting customers from a business point of view are the \textit{persuadable} and the \textit{do-not-disturb} clients. The persuadable clients provide incremental sales whereas the do-not-disturb individuals should be avoided because the marketing campaign would have a negative effect on them. Uplift models mainly aim to detect these groups of customers. Most existing approaches are adaptations of classification and regression trees \citep{radcliffe2011real, jaskowski2012uplift, guelman2012random}. Based on several experiments on real data, and although the literature suggests that tree-based methods are state-of-the-art tools for uplift \citep{soltys2015ensemble}, predicting which group a customer belongs to still lacks satisfactory solutions. One problem is that uplift is estimated using either i) the difference in conditional means (i.e. an indirect estimation defined as the difference between treated and control positive response proportions),  or ii) the transformed outcome mean \citep{athey2015machine}. In both cases, there is a harmful compromise. Either the model focuses on learning the underlying binary classification task, or it ignores it completely. This is particularly problematic for uplift trees, as the split criteria is function of the two treatment effects. Splits are likely to be placed next to extreme values because outliers of any treatment can influence the choice of a split point. In addition, successive splits tend to group together similar extreme values, introducing more variance in the prediction of uplift \citep{zhao2017practically}. As a result, the published models over-fit the training data. 

In this work, we define a unified framework with the goal of improving uplift models predictions on unseen data. We take advantage of both the conditional means and the transformed outcome in order to find a representation (using neural networks) of the data that helps in reducing generalization error. 
The main contribution of this article is to give what is, to the best of our knowledge, the first framework which allows a balance between direct estimation of the uplift and coherent prediction of the binary outcome. Our architecture is quite simple which makes it flexible. This allows us to propose an alternative to using the transformed outcome for predicting uplift. Furthermore, it opens a door to connections with the estimation of ITE in observational data settings. Our proposed method is validated on both synthetic and a realworld data-set made available to us by a Canadian insurance company. The experimental results demonstrate its advantages over state-of-the-art uplift modeling.

\section{Problem Setup}
\label{setup}

To formalize the problem, let $Y$ $\in$ $\lbrace0,1\rbrace$ be a binary response variable, $\textbf{X}=(X_{1},\ldots,X_{p})$ a vector of covariates, and $T\in\lbrace0,1\rbrace$ the treatment indicator variable. The binary variable $T$ indicates if a customer is exposed to treatment ($T=1$) or control ($T=0$). The observed outcome $Y$ can be described as 

$$Y=TY^{(1)} + (1-T)Y^{(0)},$$

where $Y^{(0)}$  and $Y^{(1)}$ are the potential outcomes under control and treatment respectively. Suppose $n$ independent customer observations $\{y_i, \textbf{x}_i, t_i\}_{i=1}^n$ are at our disposal. Although each customer $i$ is associated with two potential outcomes, only one of them can be realized as the observed outcome $y_i$. For RCTs settings, potential outcomes $(Y^{(0)},Y^{(1)})$ are independent ($\independent$) of the treatment assignment indicator $T$ conditioned on all pre-treatment characteristics $(X_1, \ldots, X_p)$, i.e.,

\begin{equation}
    \big( Y^{(0)}, Y^{(1)} \big) \independent T \mid \X. 
    \label{eq:randomization}
\end{equation}

This allows the unbiased estimation of the average treatment effect (ATE). Moreover, the \textit{propensity score} $e(\X)= \mathrm{Pr}(T=1 \mid \X)$ is constant with respect to $\X$. In our data, $e(\X) = 1/2$. An uplift model estimates the conditional average treatment effect (CATE or ITE) in various sub-populations as a function of the possible values of the covariates $\X$,

\begin{align}
    u(\X) &= \mathbb{E} ( Y^{(1)} \mid \X) - \mathbb{E} ( Y^{(0)} \mid \X) \nonumber\\
     &= \mathbb{E} ( Y \mid T=1, \X) - \mathbb{E} ( Y \mid T=0, \X) \nonumber \\
     &= \mathrm{Pr} ( Y = 1 \mid T=1, \X) - \mathrm{Pr} ( Y = 1 \mid T=0, \X) \nonumber \\
      &=m_1(\X) - m_0(\X). 
    \label{eq:uplift}
\end{align}

Uplift is naturally described as the difference between the two conditional means $m_1(\X)$ and $m_0(\X)$. Our general framework for learning uplift is depicted in Figure \ref{fig:nite}. Before formally introducing it, we must first define some quantities of interest that will be used throughout the rest of the paper. The first interesting quantity is the transformed outcome variable $Z$, which was proposed by \citep{athey2015machine} and is defined as  

\begin{equation}
    Z =  \frac{T Y}{e(\X)} - \frac{(1 - T) Y}{1 - e(\X)}.
    \label{eq:transformed_outcome}
\end{equation}

This transformation has the key property that, under the complete randomization or unconfoundness assumption \citep{rosenbaum1983central}, its expectation conditional on $\X$ is equal to $u(\X)$. Therefore, the transformed outcome is a consistent unbiased estimate of uplift. In our case, with $e(\X) = 1/2$, it simplifies to

\[ Z = \begin{cases} 
      2, & \mathrm{if}~T=1, Y=1 \\
      -2, & \mathrm{if}~T=0, Y=1 \\
      0 & \mathrm{otherwise.} 
   \end{cases}
\]

In the uplift context, we can have an interesting interpretation of $Z=2$ and $Z=-2$. Take the first case. This happens when a client is called ($T = 1$) and then buys the insurance policy ($Y = 1$). Thus, for its counterfactual (i.e., $T = 0$), there are two possibilities. Either the client would have bought the policy ($Y = 1$) and in this case would belong to the \textit{sure things} group, or else the client would not have bought the policy ($Y = 0$) in which case would in fact be a \textit{persuadable}. Thus, in the worst case scenario, individuals with $Z = 2$ are \textit{sure things}. The same logic applies for $Z = -2$ where a client could either be a \textit{sure thing} or a \textit{do-not-disturb}. For \textit{lost cause} clients, by definition $Z = 0$. This interpretation does not cover all scenarios for $Z=0$ but gives insights about ordering customers with $Y=1$. Combined with the fact that $Z$ is an unbiased estimate of uplift, this interpretation motivates us to start building our framework with a constrained version of the transformed outcome regression.

\section{General framework}
\label{methodology}

\begin{figure}[!h]
    \begin{tikzpicture}
    
    \tikzset{dist/.style={path picture= {
    \begin{scope}[x=1pt,y=10pt]
      \draw plot[domain=-6:6] (\x,{1/(1 + exp(-\x))-0.5});
    \end{scope}
    }}}
    \tikzstyle{sigmoid}=[draw,fill=gray!50,circle,minimum size=20pt,inner sep=0pt,dist]
    
    \filldraw[fill=gray!5!white, draw=black]  (0,0) rectangle (0.5,2);
    \node at (-0.25,1) {$\X$};
    \filldraw[fill=gray!10!white, draw=black] (0,-0.75) rectangle (0.5,-0.25); 
    \node at (0.25,-0.5) {$\mathbf{1}$};

    \filldraw[fill=blue!10!white, draw=black] (0,-1.5) rectangle (0.5,-1); 
    \node(treatment) at (-0.25,-1.25) {$T$};
    
    \filldraw[fill=gray!5!white, draw=black] (0,-3.75) rectangle (0.5,-1.75);
    \node at (-0.25,-2.75) {$\X$};
    \filldraw[fill=gray!10!white, draw=black] (0,-4.5) rectangle (0.5,-4); 
    \node at (0.25,-4.25) {$\mathbf{0}$};

    \filldraw[fill=gray!5!white, draw=black] (1,-0.75) rectangle (1.5,2);
    \node at (1.25,0.75) {$C_1$};
    \draw [->] (0.6,0.75) -- (0.9,0.75);
    
    \filldraw[fill=gray!5!white, draw=black] (1,-4.5) rectangle (1.5,-1.75);
    \node at (1.25,-3) {$C_1$};
    \draw [->] (0.6,-3) -- (0.9,-3);

    \filldraw[fill=gray!5!white, draw=black] (1.75,-4.5) rectangle (3.25,-1.75);
    \node at (2.5,-3) {$\boldsymbol{\theta}$};
    
    \filldraw[fill=gray!5!white, draw=black] (1.75,-0.75) rectangle (3.25,2);
    \node at (2.5,0.75) {$\boldsymbol{\theta}$};

    \filldraw[fill=green!20!white, draw=black] (3.5,-0.75) rectangle(4,2);
    \node(lp1) at (3.75,0.75) {$C_H$};
    
    \filldraw[fill=green!20!white, draw=black] (3.5,-4.5) rectangle (4,-1.75);
    \node(lp0) at (3.75,-3) {$C_H$};
    
    \node (p1)[sigmoid] at (5, 0.75) {};
    \node (p0)[sigmoid] at (5, -3) {};
    
    \filldraw (5,-1.25) circle (3pt);
    \draw [->] (0.5,-1.25) -- (5,-1.25);
    
    \draw[thick] (lp1) -- (p1) node [midway,above=-0.06cm] {};
    \draw[thick] (p1) -- (p0) node [midway,above=-0.06cm] {};
    \draw[thick] (lp0) -- (p0) node [midway,above=-0.06cm] {};
    
    \node(mu) at (4.7,-0.2) {$\mu_1$};
    \node(mu) at (4.7,-2) {$\mu_0$};
    \draw [->] (5,-1.25) -- (6.25,-1.25);
    \draw  (5.35,-3) -- (5.85,-3);
    \draw  (5.85,-3) -- (5.85,-1.5);
    \draw  (5.85,-1.5) -- (5.65,-1.5);
    \draw  (5.65,-1.5) -- (5.65,-1);
    \draw  (5.65,-1) -- (5.85,-1);
    \draw  (5.85,-1) -- (5.85,0.35);
    \draw [->] (5.85,0.35) -- (6.25,0.35);
    \draw [->] (5.35,0.75) -- (6.25,0.75);
    
    \node(mut) at (5.35,-1.1) {$\mu_T$};
    
    \filldraw[fill=gray!15!white, draw=black] (6.3,-1.6) rectangle (7.5,-0.9);
    \node at (6.9,-1.25) {$\mathcal{L}_{\boldsymbol{\theta}}(\mu_T)$};
    
    \filldraw[fill=gray!15!white, draw=black] (6.3,0) rectangle (7.9,1);
    \node at (7.07,0.5) {$\mathcal{J}_{\boldsymbol{\theta}}(\mu_0,\mu_1)$};
    
    \end{tikzpicture}
    \caption{SMITE: Neural network architecture for uplift estimation.}
    \label{fig:nite}
\end{figure}
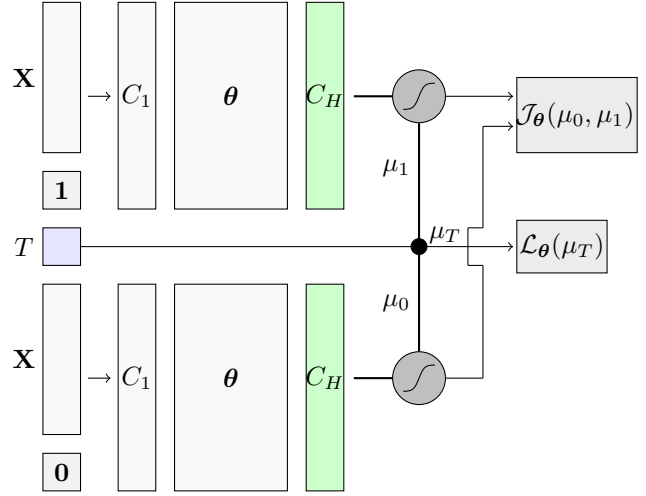

We propose a \textit{Siamese} neural network \citep{bromley1994signature} architecture called SMITE \footnote{Siamese Model for Individual Treatment Effect} which works in tandem on two different input vectors $(\X,\mathbf{1})$ and $(\X,\mathbf{0})$, and outputs the prediction of the conditional mean $m_T(\X) = Tm_1(\X) + (1-T)m_0(\X)$ as well as the one of the uplift $u(\X)$. Two input vectors are fed to the two identical neural network sub-components (i.e., with shared weights $\boldsymbol{\theta}$). The inputs contain the covariates vector $\X$ and, for the top network sub-component, the treatment variable fixed to $1$. The treatment variable is fixed to $0$ for the bottom network sub-component as shown in Figure \ref{fig:nite}. The sub-components output the predicted conditional means for treated ($\mu_1$) and for control ($\mu_0$). The predicted conditional mean $\mu_T$ is based on the actual treatment $T \in {0,1}$, i.e., $\mu_T = T\mu_1 + (1-T)\mu_0$. Then $\mu_T$ is compared to the outcome $Y$ based on binary cross-entropy (BCE) loss $\mathcal{L}$. Moreover, the difference of the predicted output values $\mu_1$ and $\mu_0$ gives direct prediction for $u(\X)$. This prediction is compared to the transformed outcome $Z$ based on the mean squared error (MSE) loss $\mathcal{J}$.

Note that an alternative to using $Z$ is easily implemented in the same framework and is explained in Section \ref{j2}. The intuition behind the SMITE architecture is based on two factors. 

\begin{enumerate}
  \item Classification models with binary outcomes are relatively easy to develop\footnote{Depending on the complexity of the data, any supervised learning algorithm used properly tends to give satisfactory results.}. However, these models fail to properly capture the ITE because they are not designed for this task.
  \item Methods that aim to directly predict the ITE fail to generalize because they focus only on ITE and do not take into account the underlying binary classification task.
\end{enumerate}

We developed the SMITE architecture with the above observations in mind. Essentially, back-propagating both losses gradients through the shared representation $C_H$ consists in fitting a {\it trade-off} between the conditional means and the transformed outcome regressions using a single model. With little fine-tuning, this representation takes advantage of both for improving uplift prediction.

\subsection{The optimization problem}\label{j1}

We want to train a model with parameters $\boldsymbol{\theta}$, by solving a constrained optimization problem. Formally, using the transformed outcome $Z$, we want to solve 

\begin{align} 
     {\boldsymbol{\hat\theta}} & = \argmin_{\boldsymbol{\theta}} ~ \mathcal{J}\Big({\mu_0}(\boldsymbol{\theta}),{\mu_1}(\boldsymbol{\theta}), Z\Big)  \\ 
     & \mathrm{subject~to,} \nonumber \\
        & \mathrm{d}\Big(\mathrm{Pr}(Y=1 \mid T, \X), {\mu_T}(\boldsymbol{\theta}) \Big) < \epsilon,
     \label{eq:objective}
\end{align}

where $\mathrm{d}(.)$ is a distance and $\epsilon > 0$. $\mu_{.}(.)$ are functions of the inputs $\X$ and should be denoted $\mu_{.}(\X)$. To simplify the notation, we prefer to omit this specification throughout the rest of the paper. The objective function $\mathcal{J}(.)$ is what we refer to as the \textit{direct} uplift loss function. The rationale behind this constrained formulation is that we want a model capable of direct prediction of the uplift $u(\X)$, i.e., $\mu_1(\boldsymbol{\hat\theta}) - \mu_0(\boldsymbol{\hat\theta})$, and coherent estimation of $m_T(\X)$, i.e., $\mu_T(\boldsymbol{\hat\theta})$.

The direct uplift loss can be defined as a distance between the predicted uplift and the transformed outcome. A natural choice is the mean squared error (MSE),

\begin{equation}
    \mathcal{J}(\boldsymbol{\theta}, Z) =   \frac{1}{n} \sum_{i=1}^{n} \Big( z_i - \{{\mu_{1i}}(\boldsymbol{\theta }) - {\mu_{0i}}(\boldsymbol{\theta})\} \Big)^2.
    \label{eq:direct_uplift_loss}
\end{equation}

A convenient choice of distance $\mathrm{d}(.)$ in our case is binary cross-entropy (BCE). Formally, we define $\mathcal{L}(\boldsymbol{\theta}, Y)$ with respect to the parameters $\boldsymbol{\theta}$ as 

\begin{equation*}
    \mathcal{L}(\boldsymbol{\theta}, Y) = - \sum_{i=1}^{n} \Big( y_i\mathrm{log} \{\mu_{t_i}(\boldsymbol{\theta})\} + (1-y_i) \mathrm{log} \{1-\mu_{t_i}(\boldsymbol{\theta})\} \Big)
\end{equation*}

where $\mu_{t_i}(\boldsymbol{\theta}) = t_i \mu_{1i}(\boldsymbol{\theta}) + (1-t_i) \mu_{0i}(\boldsymbol{\theta})$ is the model's output of observation $i$ with $T=t_i$. It is helpful to rewrite the full constrained problem (\ref{eq:objective}) in his Lagrangian form

\begin{align*}
    {\boldsymbol{\hat\theta}}_{\lambda} =  \argmin_{\boldsymbol{\theta}} \mathcal{J}(\boldsymbol{\theta}, Z)+ \lambda  \mathcal{L}(\boldsymbol{\theta}, Y)
\end{align*}

where $\lambda > 0$ can be seen as a regularization constant. However, in the SMITE framework, we use the transformation $\alpha = 1 / (1+\lambda)$. This transformation allows to interpret $\alpha \in [0,1]$ as a hyper-parameter which controls the trade-off between the two losses in the optimization problem. Hence,

\begin{align}
    {\boldsymbol{\hat\theta}}_{\alpha} = \argmin_{\boldsymbol{\theta}} (1 - \alpha) \mathcal{J}(\boldsymbol{\theta}, Z)+ \alpha  \mathcal{L}(\boldsymbol{\theta}, Y).
    \label{eq:SMITE_objective}
\end{align}

Equation (\ref{eq:SMITE_objective}) is the one of greatest interest. It encompasses exactly what motivates the SMITE framework, to jointly optimize both loss functions. 

In practice, we train the SMITE models by minimizing (\ref{eq:SMITE_objective}) using the stochastic gradient descent (SGD) optimizer with fixed learning rate. In our experiments, the {\it trade-off} constant $\alpha$ is fine-tuned using cross-validation (see Figure \ref{fig:alphaie}). Similar experiments were performed for the learning rate $\xi$.

\subsection{An alternative to the $Z$ transform} \label{j2}

\begin{figure}
    \centering
    \includegraphics[width=0.35\textwidth]{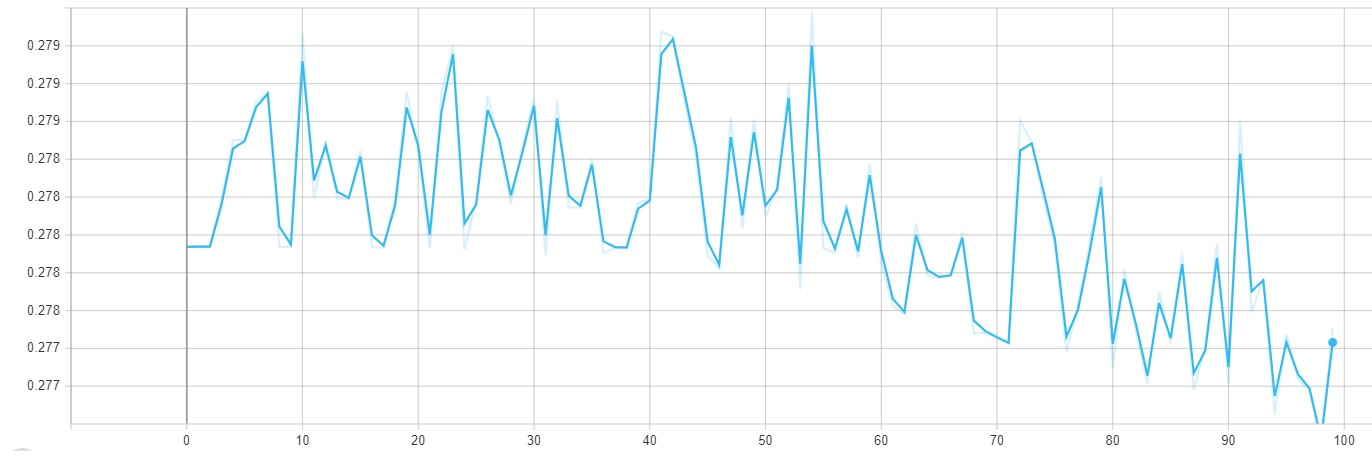}
    \includegraphics[width=0.35\textwidth]{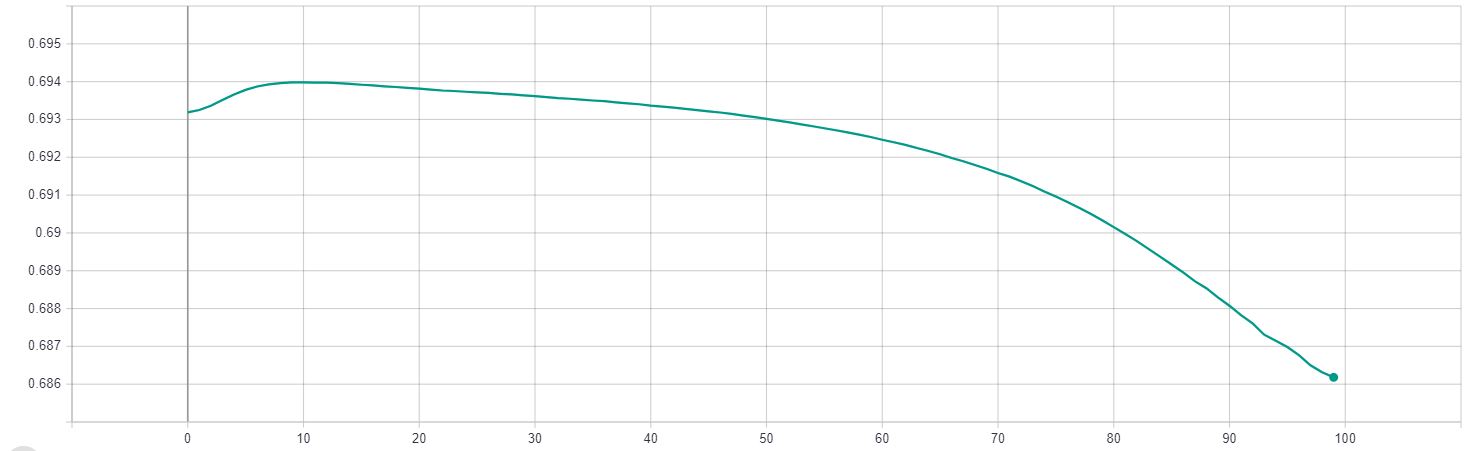}
    \caption{Uplift loss functions learning curves:  direct uplift loss $\mathcal{J}$ (top) and indirect uplift loss $\mathcal{I}$ (bottom).}
    \label{fig:instabilty}
\end{figure}

As mentioned before, the flexibility of SMITE allows us to offer alternatives to the loss functions defined in the general framework. For example, instead of using the direct uplift loss (\ref{eq:direct_uplift_loss}), one can think of the $L_1$-norm loss. Furthermore, we noticed that the learning curve associated with (\ref{eq:direct_uplift_loss}) could have erratic behavior in early stages of training (see Figure \ref{fig:instabilty}). This led us to define an alternative to the $Z$ transform which improves the performance of our uplift models.

Our intuition started with the following observation :

	{\it "Suppose a treatment has a significant positive (resp. negative) effect on a sub-sample of customers. Therefore, within the sample of customers that had a positive (resp. negative) response, we expect a higher proportion of treated customers."}

More precisely, we denote the proportion of treated customers among those that had positive outcomes by $\pi_1(\X)$. When $e(\X) = 1/2$, which is the case of our data\footnote{Although the equations are simplified based on this assumption, the development can easily be generalized for any constant $e(\X)$. Moreover, one can use an over-sampling procedure to recover the $e(\X) = 1/2$ case.}, it is easy to show that $\pi_1(\X)$ can be written as a function of the conditional means, 
\begin{align*}
    \pi_1(\X) = \frac{m_1(\X)}{m_0(\X) + m_1(\X)}.
\end{align*}

Similarly, the proportion of treated customers among those that had negative outcomes is
\begin{align*}
    \pi_0(\X) = \frac{(1-m_1(\X))}{(1-m_0(\X)) +(1- m_1(\X))}.
\end{align*}

From these definitions, we propose an alternative to the direct uplift loss. The \textit{indirect} uplift loss has the following form

\begin{equation*}
    \mathcal{I}(\boldsymbol{\theta}, 
    T) =   - \sum_{i=1}^{n} \Big( t_i\mathrm{log} \{\Pi_{y_i}(\boldsymbol{\theta})\} + (1-t_i) \mathrm{log} \{1-\Pi_{y_i}(\boldsymbol{\theta})\} \Big),
\end{equation*}

where, $\Pi_{y_i}(\boldsymbol{\theta}) = y_i \Pi_{1i}(\boldsymbol{\theta}) + (1-y_i) \Pi_{0i}(\boldsymbol{\theta})$, with
\begin{align*}
    \Pi_1(\boldsymbol{\theta}) &= \frac{\mu_1(\boldsymbol{\theta})}{\mu_0(\boldsymbol{\theta}) +\mu_1(\boldsymbol{\theta})}, \\
    \Pi_0(\boldsymbol{\theta}) &= \frac{(1-\mu_1(\boldsymbol{\theta}))}{(1-\mu_0(\boldsymbol{\theta})) +(1- \mu_1(\boldsymbol{\theta}))}.
\end{align*}

The indirect uplift loss is simply the BCE distance between a function of the predicted conditional means $\Pi(\boldsymbol{\hat\theta})$ and the actual treatment variable $T$. The associated learning curves are smoother than the ones of the direct uplift loss (see Figure \ref{fig:instabilty}). In Section \ref{sec:experiments}, we will refer to the SMITE using $\mathcal{J}$ as SMITE Transformed Outcome (TO) and the SMITE using  $\mathcal{I}$ as SMITE Indirect Estimation (IE). In this case, the optimization problem becomes

\begin{align}
    {\boldsymbol{\hat\theta}}_{\alpha} = \argmin_{\boldsymbol{\theta}} (1 - \alpha) \mathcal{I}(\boldsymbol{\theta}, T)+ \alpha  \mathcal{L}(\boldsymbol{\theta}, Y).
\end{align}

\section{Related Work}

The intuitive approach to model uplift is to build two classification models \citep{hansotia2002direct}. This consists of fitting two separated conditional probability models: one for the treated individuals, and another for the untreated individuals. The uplift is estimated as the difference between these two conditional probability models. The asset of this technique is its simplicity. However, it does not perform well in practice \citep{radcliffe2011real}. Both models focus on predicting only a one-class probability and for each model, information about the other treatment is never provided to the learning algorithm. Using a unique classification model corrects this drawback. Some traditional uplift estimation methods use the treatment assignment as a feature, adding explicit interaction terms between each covariate and the treatment indicator, and then train classification models to estimate the uplift \citep{lo2002true}. The parameters of the interaction terms measure the additional effect of each covariate because of the treatment. Several $k$-nearest neighbours \citep{cover1967nearest} based methods are also adopted for uplift estimation \citep{crump2008nonparametric, Alemi.etal-PersonalizedMedicine-2009, su2012facilitating}. The main idea is to estimate the uplift for an observation based on its neighbourhood containing at least one treated and one control observations. However, these methods quickly become computationally expensive for large datasets. Furthermore, adaptive tree-based methods such as random forests outperform the nearest neighbours approach, especially in the presence of irrelevant covariates \citep{wager2018estimation}. Therefore, most active research in uplift modeling is in the direction of classification and regression trees \citep{breiman1984classification}. State-of-the-art proposed methods view the forests as an adaptive neighborhood metric, and estimate the treatment effect at the leaf node \citep{su2009subgroup, chipman2010bart, guelman2012random}. In \cite{radcliffe1999differential, radcliffe2011real} or \cite{ hansotia2002direct, hansotia2002incremental}, modified split criteria that suited the uplift purpose were studied. The criteria used for choosing each split during the growth of the uplift trees is based on maximization of the difference in uplifts between the two child nodes. However, in practice, these approaches are prone to overfitting. The splitting criteria introduced are based solely on maximizing differences in uplifts. Therefore, splits are likely to be placed next to extreme values because outliers of any treatment can influence the choice of a split point. In addition, successive splits tend to group together similar extreme values, introducing more variance in the prediction of uplift \citep{zhao2017practically}.

Recently, representation learning approaches have been proposed for estimating individual treatment effect based on observational data. An approach in the literature emphasizes learning a covariate representation that has a balanced distribution across treatment and outcome \citep{johansson2016learning}. Some similar methods try to minimize the distribution difference between treated and control groups in the embedding space \citep{shalit2017estimating, yao2018representation}. Another approach is based on propensity score matching \citep{rosenbaum1983central} with neural networks \citep{alaa2017deep, schwab2018perfect} and \citep{shi2019adapting} use regularized neural networks to estimate the average treatment effect (ATE).

\section{Uplift performance metrics}
\label{sec:metrics}

The \textit{ground truth} is not observed in the case of individual treatment effect estimation. In order to evaluate uplift models, we refer to two known metrics in the uplift literature, the {\it Qini coefficient} and the {\it Kendall's uplift rank correlation}. The computation of both statistics requires the construction of the {\it Qini curve} \citep{radcliffe2007using}. These statistics are used as goodness-of-fit measures for uplift models.

For a given model, let $\hat{u}_{(1)} \geq \hat{u}_{(2)} \geq ... \geq \hat{u}_{(n)}$ be the sorted predicted uplifts. Let $\phi \in [0,1]$ be a given proportion and let $ N_{\phi} = \{i: \hat{u}_{i} \geq \hat{u}_{(\ceil{\phi n})} \} \subset \lbrace 1, \ldots, n \rbrace$ be the subset of observations with the $\phi n \times 100 \%$ highest predicted uplifts $\hat{u}_i$ (here $\ceil{s}$ denotes the smallest integer larger or equal to $s\in \real$). The \textit{Qini curve} is defined as a function $f$ of the fraction of population targeted $\phi$, where 
\begin{equation*}
    f(\phi) = \frac{1}{n_t}\biggl(\sum\limits_{i \in  N_{\phi}} y_i t_i - \sum\limits_{i \in  N_{\phi}} y_i (1-t_i) / \sum\limits_{i \in  N_{\phi}} (1-t_i) \biggr),
\end{equation*}
where $n_t = \sum_{i=1}^n t_i$ is the number of treated customers, with $f(0)=0$ and $f(1)$ is the average treatment effect (ATE). In other words, $f(\phi)$ represents the incremental number of positive responses for a fraction $\phi$ or targeted customers relative to the total number of targeted customers in the sample. The Qini helps visualize the performance of an uplift model, a bit like the ROC curve of a binary classification model. Following the same logic, a straight line between the points $(0,0)$ and $(1, f(1))$ defines a benchmark to compare the performance of the model to a strategy that would randomly target individuals. This means that if a proportion $\phi$ of the population is targeted, we expect to observe an uplift of $\phi f(1)$.
The {\it Qini coefficient} is defined as
\begin{equation*}
     q = \int_0^1 Q(\phi) ~\mathrm{d}\phi = \int_0^1 \{f(\phi) - \phi~f(1)\} ~\mathrm{d}\phi,
\end{equation*}
where $Q(\phi) = f(\phi) - \phi~f(1)$. This area can be numerically approximated using a Riemann method such as the trapezoid rule formula: the domain of $\phi \in [0,1]$ is partitioned into $K$ bins, or $K+1$ grid points $0=\phi_1 < \phi_2 < ... < \phi_{K+1} = 1$, to approximate $q$ by
\begin{equation}
  \hat{q} = \dfrac{1}{2} \sum_{k=1}^K (\phi_{k+1}-\phi_k)\{Q(\phi_{k+1}) + Q(\phi_{k})\}.
  \label{eq:q:hat}
\end{equation}

It is important to note, unlike the area under the ROC curve, $\hat{q}$ could be negative. This would simply mean that a strategy following the model in question does worse than random targeting. The second goodness-of-fit measure of an uplift model is defined by the similarity between the theoretical uplift percentiles (given by the model's predictions) and empirical percentiles (observed in the data) based on the model. This can be approximated by the {\it Kendall's uplift rank correlation} \citep{belba2019qbased} defined as
\begin{equation}
    \hat{\rho} = \frac{2}{K(K-1)} \sum_{i<j} \mathrm{sign}(\bar{\hat{u}}_i - \bar{\hat{u}}_j)~\mathrm{sign}(\bar{u}_i - \bar{u}_j),
    \label{eq:corr_coeff}
\end{equation}
where $\bar{\hat{u}}_k$ is the average predicted uplift in bin $k$,  $k \in {1,...,K}$, and $\bar{u}_k$ is the observed uplift in the same bin.

\section{Experiments}
\label{sec:experiments}

We conduct a simulation study to examine the performance SMITE uplift models. We compare its performance with existing uplift methods using three types of datasets: i) a real-world dataset from a marketing campaign; ii)  bootstrapped versions of the real-world dataset with synthetic outcomes generated from tree-based models, and iii) synthetic datasets generated from parametric models.

Ensemble methods are amongst the most powerful for the uplift case. A method that has proven to be very effective in estimating ITE is one based on generalized random forests \citep{athey2019generalized}. This method uses honest estimation, i.e., it does not use the same information for the partition of the covariates space and for the estimation of the treatment effects. This has the effect of reducing overfitting. Another candidate that we consider in our experiments is also based on random forests that were designed for uplift \citep{guelman2012random, jaskowski2012uplift, soltys2015ensemble}. For this method, we consider two different split criteria, one based on Kullback-Leibler divergence and one based on Euclidean distance.

\subsection{Fine tuning SMITE}
The backbone of the SMITE architecture is a $6$-layers feed forward MLP with the following input neurons sizes: $\{200,200,300,100,50,10\}$. The first two layers are linearly connected while the rest are connected through a leaky ReLu activation function. Both the depth and the width of this network were explored using a random grid search on data generated with the framework described in Section \ref{synt}. Two hyper-parameters are fine-tuned for each set of experiments: the learning rate $\xi$ and the trade-off constant $\alpha$. To find an adequate candidate for $\alpha$, we follow the following protocol: the dataset $D$ is split into two subsets ($D_T$,$D_H$) containing respectively $70\%$ and $30\%$ of $D$. $D_H$ remains unseen during any of our fine-tuning or model selection process. We then repeat $k$-times a random separation of $D_T$ into two datasets denoted ($D^{i}_{\mathrm{train}}$,$D^{i}_{\mathrm{valid}}$) with sizes $60\%$ and $40\%$ of $D_T$. 

\begin{figure}
    \centering
    \includegraphics[width=0.35\textwidth]{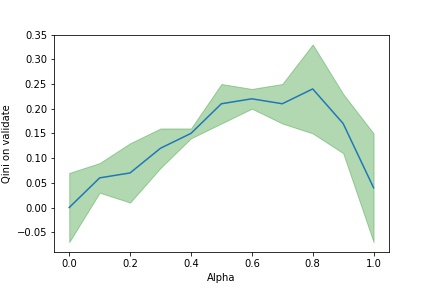}
    \includegraphics[width=0.35\textwidth]{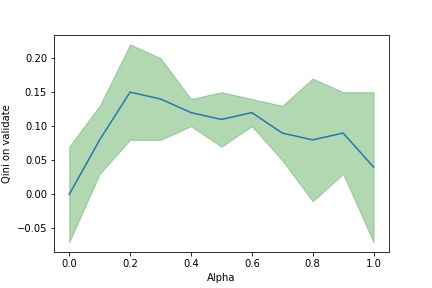}
    \caption{Fine-tuning of the trade-off constant $\alpha$ on the Canadian insurer dataset. The top curve corresponds to the (IE) loss and the bottom one to the (TO) loss. }
    \label{fig:alphaie}
\end{figure}

At a fixed learning rate of $\xi=0.03$, we fit the network on $D^{i}_{\mathrm{train}}$ for different values of $\alpha$, from $0$ to $1$ with $0.1$ increments. We measure the average Qini coefficient $\hat q$ (\ref{eq:q:hat}) obtained on $\{D^{i}_{valid}\}_{i \in [1,k]}$ and we estimated a $95\%$ confidence interval (CI) for $\hat q$. The selected $\alpha$ corresponds to the highest average as long as the $95\%$ CI lower bound remains higher than $0$. The resultant of this protocol is illustrated in Figure \ref{fig:alphaie}. Once $\alpha$ is fixed, we follow then the same process for the learning rate $\xi$.

\subsection{Data}

A data set with $n=50,000$ customers and $p=40$ covariates from an insurer is at our disposal to evaluate the performance of our methods. This company was interested in designing strategies to maximize its conversion rate. An experimental acquisition campaign was implemented for 6 months, for which half of the potential clients were randomly allocated into a treatment group and the other half into a control group. Potential clients under the treatment group were contacted using a different (and more costly) strategy. The goal of the analysis was to be able to select who among the clients portfolio should be contacted in that way going forward in order to maximize the benefits. The observed difference in sales rates between the treated group and the control shows some evidence of a positive impact of the treatment ($\mathrm{ATE}=0.55\%$).

We also evaluate the performance of SMITE using synthetic data generated using two types of mechanisms. First, we leverage the real-world data to generate semi-synthetic datasets where the outcome is generated from a known non-parametric model. We first fit an uplift random forest (RF) to a random sample $S$ of the real-world dataset. Then, we sample a bootstrap replicate of $S$ for which we generate a synthetic outcome variable based on the RF conditional means. This method is similar to the one used in \cite{belba2019qbased}. We repeat this process several times. Each bootstrapped data size is $n=10,000$ observations and $p=40$ covariates. These simulations allow us to evaluate if SMITE can recover the right representation for a model generated from a tree-based model. This also makes comparisons to random forest methods fair. 

The third simulation study is based on fully synthetic datasets generated from a parametric model inspired by the scenarios proposed by \cite{tian2014simple}. In this case, we generate datasets of sizes $n=20,000$ observations and $p=100$ covariates.

\subsection{Benchmarks and implementation}

For random forests (RF), we use the implementation from the \textbf{R} libraries \textbf{grf}  \citep{tibshirani2019R_grf}  and \textbf{uplift} \citep{guelman2014R_uplift} which were developed by the authors of the methods respectively. Implementations of these causal forests have several useful parameters that make it easy to fine tune them. In particular, \textbf{grf} provides cross-validation procedures for hyper-parameters tuning which we use to get the best possible candidates to compare our results to. We also use the \textbf{R} library \textbf{tools4uplift} \citep{belba2020R_tools4uplift} for computing the model's goodness-of-fit measures. SMITE is implemented in \textbf{Python}, so we reimplemented the Qini (\ref{eq:q:hat}) and Kendall (\ref{eq:corr_coeff}) coefficients. In total, we consider $4$ different random forests to compare to SMITE: causal forest, honest causal forest (H), RF with Kullback-Leibler (KL) and RF with Euclidean distance (ED) split criteria. 

\subsection{Experimental results}

For all models we follow the same learning process and performance metrics computations.

\subsubsection{Real data}

For the real-world dataset, a sample of $30\%$ is dedicated to out-of-sample performance measures. We do not score these customers until all models are properly fine-tuned. The remaining $70\%$ is used for fine-tuning. We repeat the experiment $30$ times. We randomly split the subset into $60\%$ for training and $40\%$ for validation. We select the model which performs best on the validation set. Using that model, we score the out-of-sample customers and compute the performance metrics. Results are presented in Table \ref{tab:realdata_perf}.

\begin{figure}[H]
    \centering
    \includegraphics[width=0.35\textwidth]{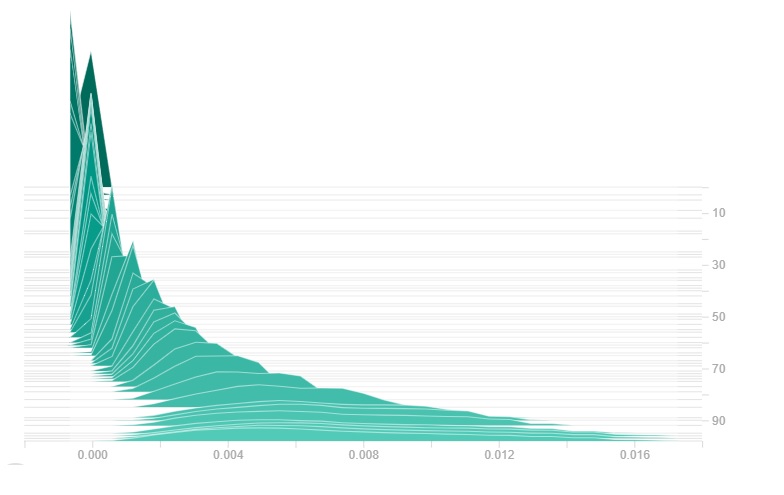}
    \includegraphics[width=0.35\textwidth]{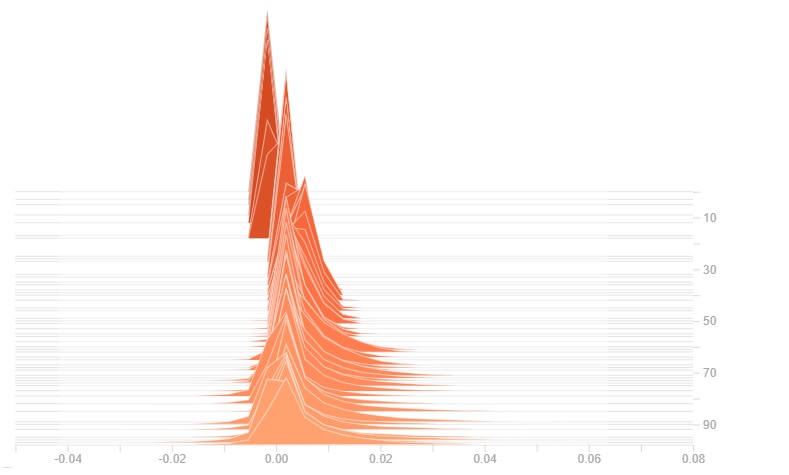}
    \caption{Examples of predicted uplift distributions on real data. We show evolution as the number of epochs increases. (top): For SMITE, individual uplifts are pushed away from zero and cover a small range within the interval $[0.000; 0.016]$ as training goes on. (bottom): A difference in conditional means neural network regression. The range covered grows as the number of epochs increases and stays more or less centered at $0$. The final interval is about $[-0.02,0.03]$. That behaviour is associated with a model that overfits .}
    \label{fig:pred_uplift_evol}
\end{figure}

We observe that the random forests tend to overfit. Indeed, based on the out-of-sample results from the hold out set, the models seem to be close to random targeting. This might be due to the small (in magnitude) overall impact of the initiative. On the other hand, we observe a good adequacy of SMITE for both TO and IE. Overall, the results of SMITE (IE) appear to outperform all other methods. It is also interesting to note that the distribution of the ITE obtained by the model SMITE (IE) are in line with what was logically expected (see Figure \ref{fig:pred_uplift_evol}: giving what appears to be a better service only slightly increases the propensity of buying.

\begin{table}[]
    \caption{Results on Canadian insurance dataset. Reported performance metrics are averaged over $30$ runs. }
    \centering
    \begin{tabular}{|p{2.9cm}||p{2.1cm}|p{1.8cm}|}
         \hline
         Model &  $\hat{q} \pm 2\mathrm{S.E.}$ & $\hat{\rho} \pm 2\mathrm{S.E.}$\\
         \hline
         \multicolumn{3}{|c|}{Training Set ($n=21,000$, $p=40$)} \\
         \hline
         Causal Forest   & $5.894 \pm 0.056$ & $0.96 \pm 0.01$\\
         Causal Forest (H) & $2.690 \pm 0.162$ & $0.97 \pm 0.01$\\
         Random Forest (ED) & $1.245 \pm 0.041$ & $0.91 \pm 0.07$\\
         Random Forest (KL) & $1.179 \pm 0.040$ & $0.91 \pm 0.025$\\
         SMITE (TO) & $0.285 \pm 0.030$ & $0.56 \pm 0.06$ \\
         SMITE (IE) & $0.322 \pm 0.034$ & $0.64 \pm 0.04$\\
         \hline
         \multicolumn{3}{|c|}{Validation Set ($n=14,000$, $p=40$)} \\
         \hline
         Causal Forest   & $0.004 \pm 0.035$ & $0.01 \pm 0.11$\\
         Causal Forest (H) & $0.035 \pm 0.063$ & $0.02 \pm 0.10$ \\
         Random Forest (ED) & $-0.079 \pm 0.04$ & $-0.18 \pm 0.08$\\
         Random Forest (KL) & $-0.077 \pm 0.05$ & $-0.09 \pm 0.11 $\\
         SMITE (TO) & $0.182 \pm  0.023$ & $0.39 \pm 0.07$\\
         SMITE (IE) & $\mathbf{0.261} \pm 0.037$ & $\mathbf{0.48} \pm 0.07$\\
         \hline
         \multicolumn{3}{|c|}{Out-of-sample ($n=15,000$, $p=40$)} \\
         \hline
         Causal Forest   & $0.051 \pm 0.039$ & $0.06 \pm 0.11$ \\
         Causal Forest (H) & $0.056 \pm 0.055$ & $0.06 \pm 0.11$\\
         Random Forest (ED) & $0.080 \pm 0.026$ & $0.05 \pm 0.08$\\
         Random Forest (KL) & $-0.002 \pm 0.03$ & $-0.06 \pm 0.09$\\
         SMITE (TO) & $0.167 \pm 0.028$ & $0.33 \pm 0.05 $\\
         SMITE (IE) & $\mathbf{0.244} \pm 0.030$ & $\mathbf{0.51}  \pm 0.08$ \\
         \hline
    \end{tabular}
    \label{tab:realdata_perf}
\end{table}

\subsubsection{Simulated data}\label{synt}

For both simulated datasets, we do not consider any out-of-sample subset. In Monte Carlo settings, we are able to generate as much data as we want. Therefore, we first fine-tuned the methods on a single synthetic dataset of same size as the one we used in the repeated simulations presented. We generate $n=10,000$ observations with $p=40$ covariates for the bootstrapped data. For the fully synthetic datasets, we report results from experiments with $n=10,000$ observations and $p=100$ covariates. We split into training and validation and compare the results averaged over $30$ runs for the models that performed best on the validation set. Results are described in Table \ref{tab:boot_perf} for the bootstrap datasets and Table \ref{tab:synthe_perf} for the synthetic datasets. For the fully synthetic data, we see that Causal Forest (H) and SMITE (TO) had similar performance on the validation set while SMITE (IE) significantly outperformed all the other methods. For the bootstrap data for which the outcomes were generated based on random forest model, it is interesting to see that SMITE (TO) and Causal Forest (H) give similar results. Moreover, once again, the SMITE (IE) model outperforms all the other models.

\begin{table}[]
    \caption{Results on bootstrap semi-synthetic data. Reported performance metrics are averaged over $30$ runs. }
    \centering
    \begin{tabular}{|p{2.9cm}||p{2.1cm}|p{1.8cm}|}
         \hline
         Model &  $\hat{q} \pm 2\mathrm{S.E.}$ & $\hat{\rho} \pm 2\mathrm{S.E.}$\\
         \hline
         \multicolumn{3}{|c|}{Training Set ($n=7,000$, $p=40$)} \\
         \hline
         Causal Forest   & $2.75 \pm 0.06 $ & $0.94 \pm 0.01$ \\
         Causal Forest (H) & $0.78 \pm 0.09 $ & $0.66 \pm  0.06$ \\
         Random Forest (ED) & $1.78 \pm 0.05$ & $0.90 \pm 0.02$ \\
         Random Forest (KL) & $1.47 \pm 0.04$ & $ 0.84 \pm 0.03$ \\
         SMITE (TO) &$0.53 \pm 0.12$  &$ 0.56 \pm 0.09$ \\
         SMITE (IE) &$0.73 \pm 0.08$  & $0.74 \pm 0.02$\\
         \hline
         \multicolumn{3}{|c|}{Validation Set ($n=3,000$, $p=40$)} \\
         \hline
         Causal Forest & $0.07 \pm 0.10 $ & $0.05 \pm 0.09$ \\
         Causal Forest (H) & $0.12 \pm 0.12$ & $0.08 \pm 0.11$ \\
         Random Forest (ED) & $0.02 \pm 0.09$ & $0.01 \pm 0.09$ \\
         Random Forest (KL) & $0.02 \pm 0.10$ & $0.04 \pm 0.09$ \\
         SMITE (TO) &$0.15 \pm 0.07$  &$0.27 \pm 0.07$  \\
         SMITE (IE) & $\bold{0.28} \pm 0.05$ &$\bold{0.34} \pm 0.06$ \\
         \hline
    \end{tabular}
    \label{tab:boot_perf}
\end{table}

\begin{table}[]
     \caption{Results on synthetic data. Reported performance metrics are averaged over $30$ runs. }
    \centering
    \begin{tabular}{|p{2.9cm}||p{2.1cm}|p{1.8cm}|}
         \hline
         Model &  $\hat{q} \pm 2\mathrm{S.E.}$ & $\hat{\rho} \pm 2\mathrm{S.E.}$\\
         \hline
         \multicolumn{3}{|c|}{Training Set ($n=14,000$, $p=100$)} \\
         \hline
         Causal Forest   & $8.90 \pm 0.10$  & $0.99 \pm 0.00$\\
         Causal Forest (H) & $4.06 \pm 0.10$ & $0.97 \pm 0.01$ \\
         Random Forest (ED) & $4.74 \pm 0.09$ & $ 0.96 \pm 0.01$ \\
         Random Forest (KL) & $4.57 \pm 0.10$ & $0.96 \pm 0.01$ \\
         SMITE (TO) & $3.98 \pm 0.40$  & $0.93 \pm 0.02$ \\
         SMITE (IE) &$3.82 \pm 0.12$  & $0.93 \pm 0.01$\\
         \hline
         \multicolumn{3}{|c|}{Validation Set ($n=6,000$, $p=100$)} \\
         \hline
         Causal Forest   & $2.09 \pm 0.17$ & $0.76 \pm 0.06$ \\
         Causal Forest (H) & $2.05 \pm 0.13$   & $0.75 \pm 0.04$ \\
         Random Forest (ED) & $1.89 \pm 0.14$ & $0.73 \pm 0.04$ \\
         Random Forest (KL) & $1.68 \pm 0.10$ & $0.72 \pm 0.05$ \\
         SMITE (TO) &  $2.31 \pm 0.22$ & $0.83 \pm 0.03$ \\
         SMITE (IE) &  $\mathbf{2.80} \pm 0.12$ & $\mathbf{0.89} \pm 0.02$\\
         \hline
    \end{tabular}
    \label{tab:synthe_perf}
\end{table}

\section{Conclusion}
\label{discussion}

In this paper we present a meaningful and intuitive neural networks architecture for the problem of uplift modeling. Our framework, which we call SMITE is flexible enough to jointly optimize both the conditional mean regression and the transformed outcome regression. As a result, the SMITE framework makes it easier to predict uplift and reduce generalization error. This flexibility also allowed us to easily introduce an alternative to the MSE loss function. This alternative indirect estimation of uplift complements our unified framework. We apply our methods to both real-world and synthetic data and we compare with what is, to the best of our knowledge, state-of-the-art methods for uplift. Our results show that the SMITE models significantly outperform random forests designed for ITE and uplift in all of our scenarios. For future work, we want to study the theoretical properties of the SMITE optimization problem as well as including more causal inference related problems such as multiple treatments and the use of observational data.

\bibliography{main}
\bibliographystyle{icml2020}

\end{document}